\documentclass[conference]{IEEEtran}
\IEEEoverridecommandlockouts
\usepackage{cite}
\usepackage{amsmath,amssymb,amsfonts}
\usepackage{graphicx}
\usepackage{textcomp}

\usepackage{algorithm}
\usepackage{algpseudocode}
\usepackage[misc,geometry]{ifsym} 
\usepackage{mathtools}
\usepackage{graphicx}
\usepackage{color}
\usepackage{multirow}
\usepackage{array}
\usepackage{booktabs}
\usepackage{url}
\usepackage{subfig}
\usepackage[misc,geometry]{ifsym}

\usepackage[table]{xcolor}
\definecolor{rcColor}{rgb}{0.96,0.93,0.93}
\definecolor{cellColor}{rgb}{0.96,0.94,0.94}

\def\BibTeX{{\rm B\kern-.05em{\sc i\kern-.025em b}\kern-.08em
    T\kern-.1667em\lower.7ex\hbox{E}\kern-.125emX}}
\begin{document}

\title{Generative Adversarial Stacked Autoencoders for Facial Pose Normalization and Emotion Recognition
}

\author{\IEEEauthorblockN{Ariel Ruiz-Garcia \textsuperscript{\Letter}}
\IEEEauthorblockA{\textit{School of Computing, Electronics and Mathematics} \\
\textit{Coventry University}\\
Coventry, UK \\ 
ariel.9arcia@gmail.com}

\and
\IEEEauthorblockN{Vasile Palade}
\IEEEauthorblockA{\textit{Research Centre for Data Science} \\
\textit{Coventry University}\\
Coventry, UK \\
vasile.palade@coventry.ac.uk}
\and

\IEEEauthorblockN{Mark Elshaw}
\IEEEauthorblockA{\textit{School of Computing, Electronics and Mathematics} \\
\textit{Coventry University}\\
Coventry, UK \\
mark.elshaw@coventry.ac.uk}
\and

\IEEEauthorblockN{Mariette Awad}
\IEEEauthorblockA{\textit{Humans and Machines Lab} \\
\textit{American University of Beirut}\\
Beirut, Lebanon \\
ma162@aub.edu.lb}


}

\maketitle

\begin{abstract}

    
    In this work, we propose a novel Generative Adversarial Stacked Autoencoder that learns to map facial expressions with up to $\pm60$ degrees to an illumination invariant facial representation of $0$ degrees. We accomplish this by using a novel convolutional layer that exploits both local and global spatial information, and a convolutional layer with a reduced number of parameters that exploits facial symmetry. Furthermore, we introduce a generative adversarial gradual greedy layer-wise learning algorithm designed to train Adversarial Autoencoders in an efficient and incremental manner. We demonstrate the efficiency of our method and report state-of-the-art performance on several facial emotion recognition corpora, including one collected in the wild. 

\end{abstract}


\section{Introduction} 

    Facial expression recognition continues to be of great interest in the machine learning community due to the many challenges it presents. Many works in the literature have proposed a variety of models that produce state-of-the-art accuracy on various facial expression corpora collected in controlled environments. However, most of these models are unable to deal with non-frontal facial expression images or with drastic changes in the environment, such as different lightning conditions. This is evident on models trained on data with nonuniform conditions collected in the wild \cite{Levi}, for which state-of-the-art is significantly lower than on datasets taken in controlled environments. This is partly due to the fact that in non-frontal faces \textemdash i.e. faces with pose greater than $0$ degrees \textemdash much of the information essential for emotion recognition is nonexistent. Moreover, the more variations in facial pose the larger the data distribution, and the more difficult for a neural network to provide good generalization due to the high dimensional search space. In addition, for real-time emotion recognition in unconstrained environments, it is difficult to obtain images without facial pose. In this work we introduce a model that explicitly addresses facial pose and illumination invariance. The proposed model can deal with faces with a facial pose of up to $\pm60$ degrees and several degrees of illumination. Our main contributions can be summarized as: 

	\begin{itemize}
    	\item a novel deep Generative Adversarial Stacked Convolutional Autoencoder model that learns to map faces with facial pose of up $\pm60$ degrees to $0$ degrees representations. 		
    	
    	\item a hybrid deep learning layer employing convolutional filters to retain spatial information and learn salient features, and fully connected units shared across the depth dimension to facilitate the reduction of facial pose. 
    	
    	\item a convolutional layer with reduced number of parameter that exploits facial symmetry and learns from only one half of the face. 
    	
    	\item a gradual greedy layer-wise algorithm for Generative Adversarial Autoencoders.
    	
    	\item an illumination and pose invariant emotion recognition classifier that produces state-of-the-art classification performance in images taken in both controlled and unconstrained environments. 
    \end{itemize}

    We also show the difference of training for pose invariance only and for pose and illumination invariance. Our model is tested on several empirical facial expression datasets as well as on one collected in the wild. 
 
\section{Related Works}
	Deep neural networks are known to be difficult to train. This was particularly true before the introduction of several regularization techniques \cite{Krizhevsky2012} in recent years, and before Batch Normalization \cite{Ioffe2015}. Before such techniques existed, pre-training models was often a preferred choice for deep models over random initialization. Autoencoders \cite{Vincent2010} are suitable for such tasks given that they are trained in an unsupervised fashion, overcoming constrains imposed by lack of labelled data. 
	
	Autoencoders learn an encoder function $f$ that maps an input image $x$ to a hidden  representation $h = f(x)$, and learn a function $g$ that maps $h$ to a reconstruction $y = g(f(x))$ where $y$ is an approximation of $x$. However, in recent works, \cite{Ruiz-Garcia2018d} it has been established that the target reconstruction does not need to be the same as the input $x$ to the autoencoder. This is supported by the theory that to be useful, an autoencoder should only learn an approximation of the target reconstruction and not an identity function that replicates it \cite{Goodfellow2016}. 
	
	In principle this establishes that the input and target vectors in an autoencoder do not need to be the same, and therefore we can learn a function that maps an input from a given distribution to a target that lies in a different distribution. Autoencoder can also be trained in a greedy layer-wise (GLW) \cite{Bengio} fashion which has proven to be more effective than joint training \cite{Ruiz-Garcia2017}. 
	
	Generative Adversarial Autoencoders \cite{Makhzani2015} are some of the latest autoencoder models that follow the trending popularity of Generative Adversarial Networks (GANs) \cite{Goodfellow}. GANs are composed of two networks: a generative model $G$ and a discriminator model $D$. Both models are trained by playing a min-max adversarial game where the discriminator model tries to determine if a given sample is from the generator or the dataset. In contrast, the generator maps samples $z$ from a prior distribution $p(z)$ and maps it to the data space. Generative Adversarial Autoencoders follow a similar approach where the generator is an autoencoder that maps an input $x$ to a latent representation $z$ that lies in an aggregate posterior distribution $q(z)$ and back to a reconstruction $y$ which is an approximation of $x$. The discriminator network in this framework attempts to determine if a sample has been drawn from a prior distribution $p(z)$ or from the latent distribution $q(z)$. 
	
	Although GANs are mainly used for data synthesis, some works have explored their use in classification \cite{Israel2017}. For emotion recognition, works employing GANs mainly focus on using GANs for data augmentations, whether in emotion recognition from speech \cite{Chatziagapi2019} or from facial expressions \cite{Yi2018,Zhu2017}. Most other works focus on generating data, but do not explore emotion recognition. Such works include multi-pose face recognition \cite{Tran2017}, \cite{Zhao2017}, or facial expression image completion \cite{Chen}.

	Some works attempt to deal with some of the common challenges in emotion recognition, such as, illumination invariance. Contemporary attempts to address illumination invariance in the domain of facial expression recognition include the use of noise injection \cite{Vincent2010}, blurring images with Gaussian filters \cite{Liu2005}, a combination of histograms, principal component analysis (PCA) and discrete cosine transforms \cite{Tosik2013}, or complex models \cite{Liu2005,Gupta2016} and very deep CNN architectures \cite{Chen2017}.
	
	To the best of the knowledge of the authors, although many works target pose invariant face recognition, no existing work focuses on pose invariant facial expression recognition.

	In this work we normalize facial pose $\varphi$, where $|\varphi| >0$ to frontal images $0$ degrees pose. We also normalize images with relative luminance $Y$ to a target luminance $\mu$. Considering that autoencoders allow us to learn a mapping from an input image to a target image that does not necessarily lie in the same distribution, they are suitable for this task. This in effect means that we are interested in imposing a distribution on the input data to produce reconstructions that resemble the desired target. Adversarial autoencoders can facilitate this task as they uniformly impose a data distribution on the code vector, i.e. the hidden representation produced by the encoder element, to generate realistic reconstructions. Moreover, adversarial autoencoders are designed to produce very realistic reconstructions. 
		
\section{Generative Adversarial Stacked Convolutional Autoencoders}
	
	\begin{figure}[!htbtp]
		\centering
		\includegraphics[width=8cm]{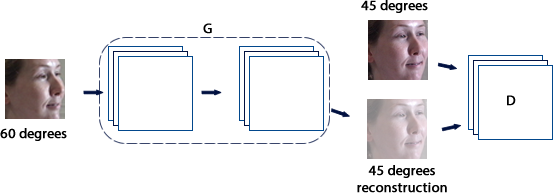}
		\caption[Facial pose reduction GASCA model]{Visualization of the first shallow autoencoder in the GASCA model.}
		\label{fig:PoseSCAE}%
	\end{figure} 

	We introduce the Generative Adversarial Stacked Convolutional Autoencoders (GASCA) framework. Just like in conventional adversarial autoencoders, in a GASCA model, the discriminator attempts to tell whether a sample comes from the training dataset or if it is a reconstruction produced by the autoencoder. 
	
	Let $x_\varphi$ be a sample from the data distribution $p_d(x_\varphi)$ and $x_\mu$ the sample from the data distribution $p_d(x_\mu)$ used as the desired target reconstruction. The autoencoder $G$ model learns to map $x_\varphi$ to a latent space $z$, note that this is not an aggregate posterior as in conventional adversarial autoencoders, and back to a reconstruction $y$ that resembles $x_\mu$ and lies in the distribution $q(y)$. The discriminator $D$ attempts to differentiate between $y$ and $x_\mu$. 
	
	In the conventional adversarial autoencoder framework \cite{Makhzani2015} a distribution $p(z)$ \textemdash often a Gaussian distribution\textemdash is imposed on $q(z)$ by estimating the divergence between $q$ and $p$. This imposition can be used to produce reconstructions with specific features. However, in this work, the objective is to produce reconstructions that are as close as possible to the desired target image $x_\mu$. Consequently, instead of imposing random noise on the hidden representation vector, the GASCA model imposes $p_d(x_\mu)$ on $q(y)$ in the following way: 
	\begin{equation}
	q(y) = \mathbb{R}_{x_\mu} q(y|x_\mu)p_d(x_\varphi)dx_\mu
	\end{equation}

	With this formulation, the discriminator model $D$ is optimized to rate samples from $p_d(x_\mu)$ with a higher probability, and samples from $q(y)$ with a low probability. Formally this is defined as: 	
	\begin{equation}
	\label{eq:GASCAd}
	\nabla_{\theta_d} \frac{1}{m} \sum_{i=1}^{m}\Big[\log D(x^{(i)}_\mu) +\log \big(1 - D(G(y^{(i)}))\big)\Big]
	\end{equation}
	where $x_\mu$ is an input image and $x_\varphi$ is the target reconstruction image. Note that since $(x_\varphi=x_\mu)$ is not necessarily true, the discriminator $D$ is not guaranteed to see the input to $G$. 
	
	The objective of the autoencoder model $G$, which in term plays the role as the generator, is to convince the discriminator model $D$ that a sample reconstruction $y$ was drawn from the data distribution $p_d(x_\mu)$ and not from $q(y)$. This optimization is done according to:  
	\begin{equation}
	\label{eq:GASCAg}
	\nabla_{\theta_g} \frac{1}{m} \sum_{i=1}^{m} \log\Big(1 - D\big(G(y^{(i)})\big)\Big)
	\end{equation}

	Furthermore, since GANs are known to be difficult to train due to their sensitivity to hyper-parameters and parameter initialization which often leads to mode collapse, the GASCA model is trained in a GLW fashion. However, since the greedy nature of GLW leads to error accumulation as individual layers are trained and stacked \cite{Ruiz-Garcia2018d}, we build on the gradual greedy layer-wise training algorithm from \cite{Ruiz-Garcia2018d} and adapt it for adversarial autoencoders. Accordingly, we introduce the GAN gradual greedy layer-wise (GANGGLW) training framework and formally define it in Algorithm \ref{tab:Gradual-GLW2}.

	\begin{algorithm}
	\caption{Given a training set $X$ and validation set $\tilde{X}$ each containing input images $x_\varphi$ and target images $x_\mu$, $m$ shallow autoencoders, an unsupervised feature learning algorithm $\mathcal{L}$ \textemdash see Algorithm \ref{tab:MiniBatch2} \textemdash which returns a trained shallow autoencoder and a discriminator model, and a fine-tuning algorithm $\mathcal{T}$ \textemdash see Algorithm \ref{tab:MiniBatchFineTuning}: train $D^1$ and $G^1$ jointly with raw data and add them to their corresponding stacks $G$ and $D$. For the remaining autoencoders and generator models: encode $X$ and $\tilde{X}$ using the encoder layers $\xi$ from the stack $G$. Create a new discriminator $D^k$ and train together with the new autoencoder $G^k$ and add them to their corresponding stacks. Fine-tune $G$ on raw pixel data. Forward propagate $x_\varphi \subset X$ through $G$ and use the resulting features, along with $x_\mu$, to fine-tune $D$ for binary classification.}
	\label{tab:Gradual-GLW2} 
	\begin{algorithmic}[1]
		\State $ [G^1,D^1] \leftarrow \mathcal{L} (G^1,D^1,X,\tilde{X})$
		\State $G \leftarrow G \circ G^1$
		\State $D \leftarrow D \circ D^1$
		\For{ $k=2$, \ldots, $m$ }
		\State $[\xi,\delta] \leftarrow D$
		\State $[X_g,\tilde{X}_g] \leftarrow \xi(X,\tilde{X})$
		\State $[G^k,D^k] \leftarrow \mathcal{L} (G^k,D^k,X_d,\tilde{X}_d)$
		\State $G \leftarrow G^{(k)} \circ G$
		\State $D \leftarrow D^{(k)} \circ D$
		\State $G \leftarrow \mathcal{T} (G, X,\tilde{X})$
		\State $ X_\varphi  \leftarrow G(x_\varphi)$
		\State $D \leftarrow \mathcal{T} (D, \{X_\varphi, x_\mu\subset X\} )$
		\EndFor
		\State \textbf{return}  $G,D$
	\end{algorithmic}
	\end{algorithm}
		
	\begin{algorithm}
		\caption{Given a training dataset $X$ with $m$ mini-batches of size $b$, an autoencoder model $G$ and discriminator model $D$ both with weight matrices $W_g$ and $W_d$, an absolute value cost function $loss$: train $G$ and $D$ jointly such that:}
		\label{tab:MiniBatch2} 
		\begin{algorithmic}[1]
			\State $V(W_d) \leftarrow \frac{2}{Nin+Nout}$
			\State $V(W_g) \leftarrow \frac{2}{Nin+Nout}$
			\For{ $k=1$, \ldots, $M$ }
			\For{ $n=1$, \ldots, $m$}
			\State  $[x_\varphi,x_\mu]_{n} \subset {1, \ldots , m} \leftarrow random(X,b)$
			\State $y_g \leftarrow predict(x_\varphi,G)$
			\State $L_g \leftarrow loss(x_\mu, y_g) $ 
			\State $G \leftarrow update(G, L_g)$
			\State $p_{\mu} \leftarrow predict(x_\mu,D)$
			\State $L_{p_d(x_\mu)} \leftarrow loss(1,p_\mu) $ 
			\State $D \leftarrow update(D, L_{p_d(x_\mu)})$
			\State $p_{y} \leftarrow predict(y_g, D)$
			\State $L_{q(y)} \leftarrow loss(0, p_y) $
			\State $ L_{adversary} = L_{p_d(x_\mu)} + L_{q(y)} $
			\State $ L_{minimax} \leftarrow  loss(1,p_y)$
			\State $ L  = L_{minimax} + L_g $
			\State $ MM_{L_g} \leftarrow  lossGrad(1,p_y)$ 
			\State $ MM_{g}\leftarrow  Grad(y_g, MM_{L-g}, D)$ 
			\State $G \leftarrow update(G, MM_g)$
			\State $Adam(L, G)$
			\State $SGD(L_{adversary}, D)$
			\EndFor
			\EndFor
			\State \textbf{return}  $G,D$
		\end{algorithmic}
	\end{algorithm}
	
	\begin{algorithm}
		\caption{Given a training dataset $X$ with $m$ mini-batches of size $b$, a validation set $\tilde{X}$ and a model $f$, train $f$ for $M$ epochs}
		\label{tab:MiniBatchFineTuning} 
		\begin{algorithmic}[1]
			\For{ $k=1$, \ldots, $M$ }
			\For{ $n=1$, \ldots, $m$}
			\State $[x,x_\mu]_{n} \subset {1, \ldots , m} \leftarrow  random(X,b)$
			\State  $y \leftarrow predict(x,f)$
			\State  $L \leftarrow loss(x_\mu, y) $
			\State  $f \leftarrow update(f, L)$
			\EndFor
			\EndFor
			\State \textbf{return} $f$
		\end{algorithmic}
	\end{algorithm}
	
	By fine-tuning $G$ and $D$ in Algorithm \ref{tab:Gradual-GLW2} we avoid error accumulation from one layer to the next and reduce the required number of fine-tuning steps for deeper layers. However, in the special case where the input and target images are significantly different, GANGGLW is not very compatible with Convolutional Neural Networks (CNNs). CNNs are designed to retain spatial information through filter kernels, whereas in GANGGLW the input and target reconstruction images are not always the same and as such spatial information often needs to be shifted or transformed, or partially ignored. For this reason, and since in our experimental set up we are trying to normalize facial pose, we introduce ConvMLP layers in the next section.

\section{ConvMLP and HalfConv layers}

	\begin{figure}[!htbtp]
		\centering
		\includegraphics[width=5cm]{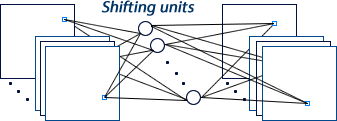}
		\caption[ConvMLP layers]{ConvMLP layers illustration. Connection weights for the \textit{shifting units} are shared between all the feature maps.}%
		\label{fig:ConvMLP}%
	\end{figure}

	One of the main advantages offered by CNNs over multilayer perceptron networks (MLPs) is their ability to self-learn a translation invariant downsampled feature vector that highlights salient features and retains spatial information through filter kernels. 
	However, CNNs are constrained to preserve the spatial structure of images and therefore are not suitable to reduce or increase facial pose:  since every output value produced by convolutional layers is the results of the dot product between a filter kernel and a small view of the input image, the pixel values can only be shifted within the space covered by the filter kernel. Normally, filter kernels tend to be small in order to capture small salient features.

	To overcome the limitations imposed by convolutional kernels and fully connected layers, and at the same time exploit the advantages offered by both, we introduce a hybrid layer that combines both approaches. The most straightforward to accomplish this is by simply placing an MLP after the convolutional layer. And, by having a smaller number of hidden units in the MLP than the number of features produced by the convolutional kernels, there would be no need for down-sampling layers such as average or max pooling or convolutional layers with a stride greater than one, which often result in the loss of important information. However, because convolutional layers normally employ a high number of convolutional kernels, this approach would require a significantly large weight matrix $W$. Accordingly, $W$ would need to have a connection weight for each feature in the feature maps produced by convolutional kernels, resulting in a large number of learnable parameters, increased computational cost, and increased training difficulty. 
	
	In contrast, the novel layer presented here, referred to as ConvMLP hereafter, shapes the resulting feature map produced by a convolution operation with a fully connected layer that is shared between all the resulting feature maps. Refer to Figure \ref{fig:ConvMLP} for a pictorial description. Given an input image $I$ and a filter kernel $K$ with $m\times n$ dimensions, and a second weight matrix $W$, the output of ConvMLP layers is defined as: 
	\begin{equation}
	\label{eq:ConvMLP}
	C(i,j) = W \big((I * K)(i,j) \big)
	\end{equation} 
	where:
	\begin{equation}
	\label{eq:ConvMLP2}
	(I * K)(i,j) = \sum_{m}\sum_{n}I(m,n)K(i-m,j-n)
	\end{equation} 
	
	Just as in empirical convolutional layers, the non-linearity is provided by a ReLU activation function, extending the above equation to:
	\begin{equation}
	\label{eq:ConvMLP3}
	y = \max(0, C(i,j))
	\end{equation}  
	
	In this formulation of ConvMLP layers, during the forward pass, the weight matrix $W$ is used to shape every feature map produced by the convolution operation and is updated only once using backpropagation. Sharing this layer across the third dimension\textemdash not taking into account the batch dimension for simplicity\textemdash its weight matrix is many orders of magnitude smaller than without weight sharing. 
	This also ensures that the $shifting$ layer learns to shift all the features highlighted in every feature plane in the same manner. Notice in Figure \ref{fig:ConvMLP} how the pixels on the second feature map are at a different location. 
	
	In addition to ConvMLP layers, and in order to support the pose invariant training approach and models presented in this work, a second convolutional layer is introduced here. This novel layer, referred to as HalfConv hereafter, exploits facial symmetry present in face images with an estimated pose of zero degrees. HalfConv layers slice the input vector vertically in half. The half containing all the facial features belonging to the left side of a face is then used as input for a convolutional layer that has half the number of parameters than an empirical convolutional layer. The resulting feature map is then simply mirrored across the $y$ axis. 
	
	When applied to face or facial expression images, HalfConv layers give up some important information on the right edge of the input image, which in effect corresponds to the features in the middle of a face. This is due to the nature of the convolution operation, which convolves a kernel across an input image, resulting in a feature plane with smaller dimensions than the input image. For this reason, HalfConv layers enforce zero padding $p$ on right side edge of the input image to allow the filter kernel to capture the features closer to the edge. Their output is then defined by: 
	\begin{equation}
	\label{eq:HalfConv}
	C(i,j) = (I * K)(i,j) = \sum_{m}\sum_{n}I(m,n)K(i-m,(j+p)-n)
	\end{equation} 
	where $p = \frac{j}{2}+1$. Then every resulting feature plane is reflected over the $y$ axis, resulting in a full image. Note that padding $p$ is enforced to avoid losing features at the edges of the image. 
	
	The main advantage offered by HalfConv layers is the reduced number of learnable parameters, which in effect results in easier and faster training. Because the only extra operation required by this layer is simply mirroring a feature vector vertically, HalfConv layers are significantly less computationally expensive than empirical convolutional layers. Furthermore, because this layer only deals with frontal faces, there is no need to employ any shifting neurons. Note that these layers are only suitable for cases where symmetry is existent in the input image or is desired in the resulting feature plane. Therefore, in the GASCA model, these layers are only used when $\alpha = 0$.

    \section{Unsupervised Feature Learning}
    	\begin{figure}[!htbtp]
    	\centering
    	\subfloat {{\includegraphics[width=1.55cm]{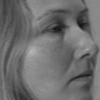} }}%
    	\subfloat{{\includegraphics[width=1.55cm]{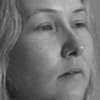} }}%
    	\subfloat{{\includegraphics[width=1.55cm]{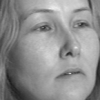} }}%
    	\subfloat{{\includegraphics[width=1.55cm]{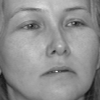} }}%
    	\subfloat{{\includegraphics[width=1.55cm]{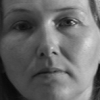} }}%
    	\\
    	\subfloat {{\includegraphics[width=1.55cm]{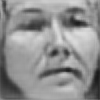} }}%
    	\subfloat{{\includegraphics[width=1.55cm]{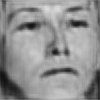} }}%
    	\subfloat{{\includegraphics[width=1.55cm]{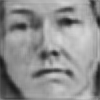} }}%
    	\subfloat{{\includegraphics[width=1.55cm]{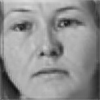} }}%
    	\subfloat{{\includegraphics[width=1.55cm]{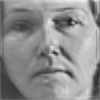} }}%
    	\\
    	\caption[GASCA$_2$ reconstructions as $0$ degrees]{Top row: input images $x_\varphi$ to the GASCA$_{2}$ model with estimated facial poses at $+60,+45,+30,+15,0$ degrees. Bottom row: corresponding reconstructions $y$ produced by the GASCA$_2$ model with an estimated pose at $\sim0$ degrees.}%
    	\label{fig:gasca_reconstructions}%
    \end{figure}

	We train two Generative Adversarial Stacked Convolutional Autoencoders, one to normalize facial pose (GASCA$_1$) and another to normalize facial pose and illumination (GASCA$_2$). We train both models using the GANGGLW training method introduced earlier. We train each shallow autoencoder to gradually reduce facial pose, or keep it the same if it is already smaller than the desired target. This process is repeated until reaching a facial pose of $0$ degrees. Effectively, the search space for the upper layers is greater than that of the deepest layer, which only has to learn one facial pose of $0$ degrees.

	For the GASCA$_2$ model we incorporate illumination invariance normalization training by taking images with disproportionate degrees of illumination and reconstructing them as images with good illumination: good illumination is determined by their relative luminance $Y$ as done by \cite{Ruiz-Garcia2018d}.
	
	We employ the MultiPie dataset \cite{Gross2008} to train these models given that it contains facial images with multi-pose and multi-illumination. 
	
	The training dataset is built according to: 
	\begin{equation}
	\label{eq:target}
		\ x_\mu =  
		\begin{cases}
		x_{\varphi-d}  \quad,\quad & \mbox{if}\quad 0 < \alpha < \varphi \\
		x_{\varphi+d}  \quad,\quad & \mbox{if}\quad \varphi < \alpha < 0  \\
		x_{\varphi}    \quad\quad,\quad & \mbox{if}\quad |\varphi| \leq |\alpha| \\
		\end{cases}
		\;
	\end{equation}
	where $-60 <= \varphi <= 60$,  $\alpha$ denotes the desired target pose, $ \alpha$ $ \in \{0,\pm15,\pm30,\pm45\}$, and $d$ denotes the change in pose by degrees: $15$ degrees in this work. Each subset is further split into $70\%$ training and $30\%$ validation subsets.

	For the GASCA$_2$ model, since every image in the MultiPie dataset has $19$ copies with different levels of relative luminance, we measure their cumulative relative luminance and pick the image closest to the mean as the target reconstruction for all others. 
	
	Since $ \alpha$ $ \in \{0,\pm15,\pm30,\pm45\}$, we design the GASCA models with three ConvMLP layers and one HalfConv Layer for the encoder element. ConvMLP layers use $5\times5, 3\times3, 3\times3$ filter kernels and 100 hidden units for the $shifting$ units. The decoder element only uses deconvolutional layers to force the encoder to learn a downsampled pose invariant hidden representation, since we fine-tune it later on to do classification. Accordingly, every shallow autoencoder is trained on a single target pose. For instance, subset $A_1$ contains all the images with $\{0,\pm15\}$ degrees. $A_2$ contains all images at angles $\{0,\pm15,\pm30\}$, thus $A_1\cap A_2$,

	Every shallow autoencoder in the GASCA models is trained for $100$ and fine-tuned for $20$ epochs. $G$ is optimized using ADAM \cite{Kingma2015a}, whereas $D$ employs SGD with Nesterov momentum. The initial learning rates for each individual shallow autoencoder in $G$ were set to $\lambda \in \{0.1, 0.3, 0.5, 0.7, 0.75\}$. Since $D$ learns faster than $G$, the shallow autoencoders employ smaller learning rates: $\lambda \in \{0.01, 0.03, 0.5, 0.07\}$. During fine-tuning, the stacks $G$ and $D$ use a learning rate of $0.001$. This combination of hyper-parameters provided the best results for both models. 
	
\section{Pose Invariant Reconstruction Results}
	
	The novel pose invariant Generative Adversarial Stacked Convolutional Autoencoder models proposed in this work are trained to gradually reduce facial pose using GANGGLW training. 
	As it can be observed in Figure \ref{fig:gasca_reconstructions}, the pose and illumination invariant GASCA$_2$ model manages to reduce facial pose in facial images with an estimated pose of up to $\pm60$ degrees. It also produces reconstructions with similar illumination.   
	
	It can also be observed that on the images with pose of $\pm60$ degrees half of the face is not visible, yet the pose invariant model manages to fill in the missing information, and more importantly keeps the shape of facial shapes which are important for emotion recognition: eyes, eyebrows, mouth, nose, cheeks, among others. Nonetheless, the greater the pose in $x_\varphi$ the poorer the quality of the reconstruction $y$. This is justified by (i) the fact that the model has to compensate for missing information, (ii) the fact that only one layer is trained specifically to deal with that particular facial pose, (iii) the smaller the pose the more the images get seen by every layer in $G$ during training, and (iv) increased network depth. 
	
	If the shallow autoencoder at step $k=1$ fails to learn a pose invariant feature vector, the shallow autoencoder at step $k=2$ will struggle even more to learn a pose invariant feature vector, and so forth. GANGGLW greatly helps to address this issue by allowing inter-layer fine-tuning, which helps strengthen the weight connections between $D^k$ and $D^{k+1}$. 
	 
	One of the main remarks observed in the reconstructions is that although these retain all the important salient features, they are visually different than the input images. These reconstructions could be improved by unsupervised fine-tuning of $G$ for a significantly longer number of epochs. Likewise, secondary methods such as super resolution CNNs \cite{Shi2016} could be used to improve the visual quality of the reconstructed images. However, because the objective of this research is to only learn a pose invariant feature vector $z$ that can be used for emotion recognition, the quality or resolution of the reconstructions is trivial.  
	
 	One of the main advantages offered by ConvMLP layers is that the number of $shifting$ neurons can be adjusted as needed. In the GASCA models, every ConvMLP layer only employs $100$, which are enough to reposition facial features and eventually reduce facial pose. Another advantage offered by ConvMLP layers is that they can be used for dimensionality reduction by mapping a feature plane to a smaller feature plane. Although, this is not evaluated in this research. 	
	
	As illustrated in Figure \ref{fig:gasca_reconstructions}, the reconstructed images also do not have a horizontal line diving the face in two, as it would be expected due to the use of HalfConv layers. When visualizing the feature planes produced by these layers, the line is somewhat visible. However, because in the final stack $G$ this layer is followed by all the layers in the decoder stack of $G$, and since the line is not visible in the target reconstruction images, it vanishes during fine-tuning. 
	
	We did not notice significant differences between the reconstructions of both GASCA$_1$ and GASCA$_2$ models. However, the reconstruction loss for the latter was marginally smaller and as seen in the next section it generalizes better. 
	
\section{Pose and Illumination Invariant Emotion Recognition }
	
	\bigskip
    \begin{table*}[!htbtp]
	\begin{center} 
		\caption[Pose invariant CNN on KDEF]{(left) Classification performance ($96.810\%$) of the CNN$_1$ model on the KDEF corpus pretrained for pose invariance. (right) Classification performance ($98.070\%$) of the CNN$_{2a}$ model on the KDEF corpus pretrained for pose and illumination invariance.}
		\renewcommand{\arraystretch}{0.8}
		\setlength{\tabcolsep}{4pt}
		\begin{tabular}{ l|lllllll   ll|lllllll }
			\hline\noalign{\smallskip}
			$  $ & $ A $ & $ D $ & $ F $ & $ H $ & $ N $ & $ Sa $ & $ Su $ &	$  $ & & $ A $ & $ D $ & $ F $ & $ H $ & $ N $ & $ Sa $ & $ Su $ \\
			\noalign{\smallskip}\toprule\noalign{\smallskip}
			$ A $ &\cellcolor{cellColor}94.44 & 1.59 & 1.59 & 0.00 & 0.79 & 1.59 & 0.00  & & $ $ &\cellcolor{cellColor}96.83 & 0.79 & 1.59 & 0 & 0.00 & 0.79 & 0.00\\
			$ D $ & 0.00 &\cellcolor{cellColor}97.60 & 0.00 & 0.00 & 0.00 & 2.40 & 0.00 & & $ $  & 0.00 &\cellcolor{cellColor}97.60 & 0.00 & 0.00 & 0.00 & 2.40 & 0.00  \\	 
			$ F $ & 000 & 0.79 &\cellcolor{cellColor}89.68 & 0.79 & 0.00 & 3.97 & 4.76  &&$ $& 000 & 0.79 &\cellcolor{cellColor}93.65 & 0.79 & 0.00 & 2.38 & 2.38  \\
			$ H $ & 0.00 & 0.00 & 0.00 &\cellcolor{cellColor}100.00 & 0.00 & 0.00 & 0.00 & & $ $  & 0.00 & 0.00 & 0.00 &\cellcolor{cellColor}100.00 & 0.00 & 0.00 & 0.00  \\
			$ N $ & 0.00 & 0.00 & 0.00 & 0.00 &\cellcolor{cellColor}100.00 & 0.00 & 0.00& & $ $  & 0.00 & 0.00 & 0.00 & 0.00 &\cellcolor{cellColor}100.00 & 0.00 & 0.00 \\
			$ Sa$ & 0.79 & 0.79 & 0.00 & 0.00 & 0.00 &\cellcolor{cellColor}98.41 & 0.00& & $ $  & 0.79 & 0.79 & 0.00 & 0.00 & 0.00 &\cellcolor{cellColor}98.41 & 0.00 \\
			$ Su$ &  0.00 & 0.00 & 2.42 & 0.00 & 0.00 & 0.00 &\cellcolor{cellColor} 97.58 & & $ $  &  0.00 & 0.00 & 0.00 & 0.00 & 0.00 & 0.00 &\cellcolor{cellColor}100.00  \\
			\noalign{\smallskip}\bottomrule	
		\end{tabular}
		\label{tab:GASCAclass1}
    	\end{center}
    \end{table*}

    Once a GASCA model is trained and fine-tuned for reconstruction, it can be used as a generic feature extractor for facial expression images. However, it can only provide feature vectors that are pose and illumination invariant but that do not necessarily discriminate between different emotions. Therefore, we fine-tune the encoder element of $G$ for classification. We discard both the discriminator model $D$ along with the decoder element $g_D$ of the generator $D$ and attach a classification layer to the encoder. 
 
	The pose invariant GASCA$_1$ model is used to initialize a classifier model, CNN$_1$, which is fine-tuned and tested on the KDEF corpus \cite{Lundqvist1998}. This dataset contains frontal and images at $\pm45$ degrees. No other publicly available datasets with multiple poses have facial expression labels. 

	The pose and illumination invariant GASCA$_2$ model is used to initialize a second classifier, CNN$_{2a}$. This model is also fine-tuned and tested on the KDEF corpus. 
    
    \begin{table}[!htbtp]
    	\begin{center} 
    		\caption[Pose and illumination invariant CNN on NAOFaces]{Classification performance ($81.36\%$) of the CNN$_{2b}$ model on the NAOFaces corpus.}
    		\renewcommand{\arraystretch}{0.8}
    		\setlength{\tabcolsep}{4pt}
    		\begin{tabular}{ l|lllllll }
    			\hline\noalign{\smallskip}
    			$  $ & $ A $ & $ D $ & $ F $ & $ H $ & $ N $ & $ Sa $ & $ Su $ \\
    			\noalign{\smallskip}\toprule\noalign{\smallskip}
    			$ A $ &\cellcolor{cellColor}92.86 &	7.14 & 0.00  & 0.00 & 0.00 & 0.00 &	0.00 \\
    			$ D $ & 8.33 & \cellcolor{cellColor}75.00 & 8.33  & 0.00 & 0.00 & 8.33 & 0.00 \\
    			$ F $ & 9.09 & 0.00  & \cellcolor{cellColor}81.81 & 0.00 & 0.00 & 0.00 & 9.09 \\
    			$ H $ & 0.00 & 0.00  & 0.00  & \cellcolor{cellColor}100.00 & 0.00 & 0.00 & 0.00\\
    			$ N $ & 3.85 & 0.00  & 3.85	 & 15.38 & \cellcolor{cellColor}57.69 & 11.54 & 7.69\\
    			$ Sa$ & 9.09 & 0.00	 & 18.18 & 0.00	& 0.00 & \cellcolor{cellColor}72.72 & 0.00\\
    			$ Su$ & 0.00 & 0.00	 & 10.53 & 0.00	& 0.00 & 0.00 & \cellcolor{cellColor}89.47\\
    			\noalign{\smallskip}\bottomrule	
    		\end{tabular}
    		\label{tab:GASCAclass3}
    	\end{center}
    \end{table}
    
	In addition, in an attempt to test the robustness of our proposed methodology, we use  GASCA$_2$ to initialize a third model CNN$_{2b}$. However, due to the lack of publicly available data taken in realistic environments with multi-pose and varying illumination, as well as labels for the emotions being expressed, we build a large dataset composed of the CK+ \cite{Lucey}, JAFFE \cite{Lyons1998a}, KDEF \cite{Lundqvist1998}, and FEEDTUM \cite{Wallhoff2006a} corpora. We refer to this corpus as combined facial expressions (CFE). Note that because, as later discussed, we obtain over $99.6\%$ on this corpus, this model is evaluated on completely novel data: the entire NAOFaces corpus \cite{Ruiz-Garcia2018b}. This set has total of $196$ images collected in unconstrained environments. Participants were $28$ 21 males and 7 females between ages 18 and 55 from at least five different ethnic backgrounds.

	As opposed to empirical CNN classifier models which employ a fully connected layer after the last convolutional layer, the classifiers in this work map the resulting feature planes produced by the last convolutional layer, which is a HalfConv layer, directly to an output SoftMax layer for classification, as done in \cite{He2015a}. 
	
	The CNN${_1}$ and CNN$_{2a}$ models are fine-tuned for $10$ epochs and, because the CFE corpus has more images, CNN$_{2b}$ is only fine-tuned for two epochs. Since the stacked autoencoders are optimized using ADAM, all classifiers are fine-tuned also using ADAM and a learning rate of $0.01$. Using a different optimizer like SGD for fine-tuning would lead to the gradients changing drastically and require a longer fine-tuning process.

\section{Pose Invariant Emotion Recognition Results}
 	
		As it can be observed in Table \ref{tab:GASCAclass1}, the pose invariant model, CNN$_1$, obtains a classification performance of $96.81\%$. In contrast, the pose invariant model that also incorporated illumination invariance obtains a state-of-the-art classification rate of $98.07\%$. The main differences in performance are observed for classes: surprise, Fear, and Angry, whereas both CNN$_1$ and CNN$_{2a}$ obtained the same classification accuracy for the remaining classes. Because both models are trained using a relatively similar approach, it is hypothesized that these discrepancies in classification performance are due to these three classes containing more images with varying image luminance, thus the pose and illumination invariant model is able to generalize better.

		The CNN$_{2b}$ model is evaluated on the NAOFaces corpus and achieves $81.36\%$ accuracy. This is significantly lower than the performance of the other models on the KDEF corpus. We attribute this lower performance to the fact that the NAOFaces contains images that are substantially more difficult, i.e. people with glasses, at different poses, and different ethnicity. Moreover, this model was not fine-tuned on any images from this corpus. This theory is further supported by the fact that when we split the CFE corpus 80$\%$ training and $20\%$ testing, we obtain 99$\%$ on the test set.

		One important observation in Table \ref{tab:GASCAclass3} is that, when looking at the missclassified images for a given class, on average $40\%$ of them are frontal images, i.e. images with zero degrees pose, and the remaining $60\%$ are those with a pose. However, because the ratio of images with a facial pose is 2:1 compared to those without one. This means that on average, more images without facial pose are missclassified. These results and observations are of great importance given that they support the pose invariant pretraining approach presented in this work.  Another observation is that not a single image from the other classes was confused with Neutral. This particular score is significant taking into account that all emotions derive from a neutral state, often resulting in low precision scores.

		Despite the good performance offered by the CNN$_{2a}$ on the NAOFaces corpus, the classification performance offered by this model is not ideal. This is attributed to one major factor: cultural differences. Because the model was trained solely on images from Caucasian people, the model has never learned to adjust to cultural difference. The NAOFaces corpus contains images of people from at least five different backgrounds including: Asian, Arab, Black, Irish and Hispanic, among others unrevealed ones. In effect, because people from different ethnic backgrounds express emotions differently \cite{Hewahi2012}, the classifier should be trained with images of participants from a wide range of ethnic backgrounds and cultures. 

\section{Comparison Against State-Of-The-Art}

	We now compare our methodology to contemporary state-of-the-art methods on the KDEF corpus. Due to the lack of contemporary work designed explicitly for pose invariant emotion recognition, the methods proposed in this work are compared against one of the most common and state-of-the-art classifiers: a ResNet \cite{He2015a}. 
	Accordingly, a ResNet-34, i.e. with $34$ parametrised layers, is trained using SGD, a momentum of $0.9$ and learning rate of $0.1$. This model is trained for $100$ on the training subset of the KDEF corpus and achieves an accuracy rate of $87.472\%$ on the test subset, as illustrated in Table \ref{tab:GASCAResnet}. Note that even though the authors of \cite{Ruiz-Garcia2017} report $92.52\%$ on the KDEF corpus, those results are only reported on frontal faces without facial pose. On the contrary, all the models in this section are evaluated on images with multiple poses, hence the marginally lower performance of the ResNet model. 
	 
	 As seen in Table \ref{tab:GASCAResnet}, the pose and illumination invariance model, CNN$_{2a}$ outperforms the state-of-the-art classifier ResNet-34 model by over $10\%$. Similarly, it outperforms CNN$_1$ marginally, supporting the pose and illumination invariant training approach. The pose invariant GASCA models also have an exponentially smaller number of parameters compared to the ResNet-34 model.  
	 
	  \begin{table}
	 	\begin{center} 
	 		\caption[Pose invariant CNN vs state-of-the-art]{Classification performance comparison on the KDEF corpus: ResNet-34 \textemdash state-of-the-art classifier; CNN$_1$ \textemdash pose invariant classifier proposed; CNN$_{2a}$ pose and illumination invariant classifier proposed. }
	 		\renewcommand{\arraystretch}{0.8}
	 		\setlength{\tabcolsep}{4pt}
	 		\begin{tabular}{ c|ccc }
	 			\hline\noalign{\smallskip}
	 			$   $ & $Resnet34$ &$CNN_1$&$CNN_{2a}$\\
	 			\hline\noalign{\smallskip}
	 			$ A $ & 84.127\% & 94.444\% & 96.825\% \\
	 			$ D $ & 85.600\% & 97.600\% & 97.600\% \\
	 			$ F $ & 73.810\% & 89.683\% & 93.651\% \\
	 			$ H $ & 98.413\% & 100.000\%& 100.000\% \\
	 			$ N $ & 90.400\% & 100.000\%& 100.000\% \\
	 			$ Sa $ & 84.921\% & 98.413\% & 98.413\% \\
	 			$ Su$ & 95.161\% & 97.581\% & 100.000\% \\
	 			\bottomrule 
	 			$Total $ & 87.472\% & 96.810\% & 98.070\% \\
	 		\end{tabular}
	 		\label{tab:GASCAResnet}
	 	\end{center}
	 \end{table}

	 The novelty of this work also arises from combining greedy layer-wise training with adversarial learning. Generative Adversarial Autoencoders are trained jointly as opposed to layer-wise. They impose a random distribution $p(z)$ on the distribution $q(z)$ produced by the encoder element of $G$, and use the resulting aggregate posterior distribution is mapped to reconstruction $y$. The discriminator $D$ tries to guess if the sample was drawn from $q(z)$ or $p(z)$. The GASCA models do not use a random distribution and instead use the reconstruction $y$ produced by forward propagating $x_\varphi$ through $G$, along with the target image $x_\mu$ as input for the discriminator. The generator $G$ is optimized to reduce the distance between $y$ and $x_\mu$. By fine-tuning the stacks $G$ and $D$ at every step $k$, both models become better at their respective job. By improving the ability of $D$ to differentiate between $y$ and $x_\mu$, $G$ is forced to produce remarkable reconstructions and learn an encoder function that produces downsampled pose invariant feature vectors. 
	  	 
	 In terms of work on pose reductions, a similar model was proposed by \cite{Kan2014}. However, the authors focused on face detection and their model does not make use of Convolutional Autoencoders and instead uses MLPs, which are prone to overfitting when applied to this problem. Furthermore, because their model does not take into account spatial information, it is unable to retain salient features that are essential for emotion recognition. Whereas the GASCA models are able to retain facial features, or compensate for missing information when this is not present in the image. Additionally, the GASCA$_{2a}$ model also takes into account illumination and produces an illumination and pose invariant feature vector.

\section{Conclusions and Future Directions}

	This work has introduced a novel pose and illumination invariant facial expression recognition model. A CNN classifier is pretrained as a Generative Adversarial Stacked Convolutional Autoencoder in a gradual greedy layer-wise semi-supervised fashion. The GASCA model learns to map an input image containing a face, with an estimate pose $\varphi$, to a hidden representation $z$ with an estimated pose of 0 degrees. Once the GASCA model is trained, the encoder element is used to initialize a CNN model which is fine-tuned for classification.
	
	The outstanding performance of the GASCA models is derived from four concepts: (i) our GANGGLW training method (ii) the ConvMLP layers with $shifting$ neurons, (iii) the HalfConv layers which take exploit of facial symmetry, and (iv) multi-pose facial expressions data. Our pose and illumination invariant method produces state-of-the-art classification performance on multi-pose facial expression corpora. Moreover, the GASCA model produces reconstruction with very small errors and is able to generalize on unseen data. 
	
	The success of the pose invariant models is in part due to ConvMLP layers, which learn salient features and shift them as needed to reduce facial pose. HalfConv layers also play an important role as they reduce the number of learning parameters. HalfConv layers were inspired by the model presented by\cite{Ruiz-Garcia2016a}, which splits the input images in half to simplify feature learning. 


	To the best of the authors' knowledge, this is the first approach that combines a greedy layer-wise training method with adversarial learning. This is also the first approach to solely focus on pose and illumination invariant emotion recognition. 
	Future work will look at exploiting the ability of our model to generate new data in order to deal with scenarios where lack of multi-pose labeled exists.


\renewcommand{\refname}{List of References}
\bibliographystyle{ieeetr} 
\bibliography{conference_101719} 

\begin{thebibliography}{10}

\bibitem{Levi}
G.~Levi and T.~Hassner, ``{Emotion Recognition in the Wild via Convolutional
  Neural Networks and Mapped Binary Patterns},'' in {\em Proceedings of the
  2015 ACM on International Conference on Multimodal Interaction - ICMI '15},
  (New York, New York, USA), pp.~503--510, ACM Press, 2015.

\bibitem{Krizhevsky2012}
A.~Krizhevsky, L.~Sutskever, and G.~E. Hinton, ``{ImageNet Classification with
  Deep Convolutional Neural Networks},'' {\em NIPS}, pp.~1106--1114, 2012.

\bibitem{Ioffe2015}
S.~Ioffe and C.~Szegedy, ``{Batch Normalization: Accelerating Deep Network
  Training by Reducing Internal Covariate Shift},'' feb 2015.

\bibitem{Vincent2010}
P.~{Vincent PASCALVINCENT} and H.~{Larochelle LAROCHEH}, ``{Stacked Denoising
  Autoencoders: Learning Useful Representations in a Deep Network with a Local
  Denoising Criterion Pierre-Antoine Manzagol},'' {\em Journal of Machine
  Learning Research}, vol.~11, pp.~3371--3408, 2010.

\bibitem{Ruiz-Garcia2018d}
A.~Ruiz-Garcia, V.~Palade, M.~Elshaw, and I.~Almakky, ``{Deep Learning for
  Illumination Invariant Facial Expression Recognition},'' in {\em Proceedings
  of the International Joint Conference on Neural Networks}, (Rio de Janeiro),
  IEEE, 2018.

\bibitem{Goodfellow2016}
I.~Goodfellow, {Bengio, Yoshua}, and A.~Courville, {\em {Deep Learning}}.
\newblock MIT Press, 2016.

\bibitem{Bengio}
Y.~Bengio, P.~Lamblin, D.~Popovici, and H.~Larochelle, ``{Greedy Layer-Wise
  Training of Deep Networks},'' tech. rep.

\bibitem{Ruiz-Garcia2017}
A.~Ruiz-Garcia, M.~Elshaw, A.~Altahhan, and V.~Palade, ``{Stacked deep
  convolutional auto-encoders for emotion recognition from facial
  expressions},'' in {\em Proceedings of the International Joint Conference on
  Neural Networks}, vol.~2017-May, pp.~1586--1593, IEEE, may 2017.

\bibitem{Makhzani2015}
A.~Makhzani, J.~Shlens, N.~Jaitly, I.~Goodfellow, and B.~Frey, ``{Adversarial
  Autoencoders},'' 2015.

\bibitem{Goodfellow}
I.~Goodfellow, J.~Pouget-Abadie, M.~M.~A. in~neural {\ldots}, and U.~2014,
  ``{Generative adversarial nets},'' {\em Advances in Neural Information
  Processing Systems 27}, pp.~2672--2680, 2014.

\bibitem{Israel2017}
S.~A. Israel, J.~Goldstein, J.~S. Klein, J.~Talamonti, F.~Tanner, S.~Zabel,
  P.~A. Sallee, and L.~McCoy, ``{Generative Adversarial Networks for
  Classification},'' in {\em 2017 IEEE Applied Imagery Pattern Recognition
  Workshop (AIPR)}, pp.~1--4, IEEE, oct 2017.

\bibitem{Chatziagapi2019}
A.~Chatziagapi, G.~Paraskevopoulos, D.~Sgouropoulos, G.~Pantazopoulos,
  M.~Nikandrou, T.~Giannakopoulos, A.~Katsamanis, A.~Potamianos, and
  S.~Narayanan, ``{Data Augmentation using GANs for Speech Emotion
  Recognition},'' 2019.

\bibitem{Yi2018}
W.~Yi, Y.~Sun, and S.~He, ``{Data Augmentation Using Conditional GANs for
  Facial Emotion Recognition},'' in {\em Progress in Electromagnetics Research
  Symposium}, vol.~2018-Augus, pp.~710--714, Institute of Electrical and
  Electronics Engineers Inc., dec 2018.

\bibitem{Zhu2017}
X.~Zhu, Y.~Liu, Z.~Qin, and J.~Li, ``{Data Augmentation in Emotion
  Classification Using Generative Adversarial Networks},'' nov 2017.

\bibitem{Tran2017}
L.~Tran, X.~Yin, and X.~Liu, ``{Disentangled representation learning GAN for
  pose-invariant face recognition},'' in {\em Proceedings - 30th IEEE
  Conference on Computer Vision and Pattern Recognition, CVPR 2017},
  vol.~2017-Janua, pp.~1283--1292, 2017.

\bibitem{Zhao2017}
J.~Zhao, L.~Xiong, K.~Jayashree, J.~Li, F.~Zhao, Z.~Wang, S.~Pranata, S.~Shen,
  S.~Yan, and J.~Feng, ``{Dual-Agent GANs for Photorealistic and Identity
  Preserving Profile Face Synthesis},'' {\em Nips 2017}, no.~15, pp.~1--11,
  2017.

\bibitem{Chen}
J.~Chen, J.~Konrad, and P.~Ishwar, ``{VGAN-Based Image Representation Learning
  for Privacy-Preserving Facial Expression Recognition},'' tech. rep.

\bibitem{Liu2005}
D.~H. Liu, K.~M. Lam, and L.~S. Shen, ``{Illumination invariant face
  recognition},'' {\em Pattern Recognition}, vol.~38, pp.~1705--1716, oct 2005.

\bibitem{Tosik2013}
C.~Tosik, A.~Eleyan, and M.~Salman, ``{Illumination invariant face recognition
  system},'' in {\em 2013 21st Signal Processing and Communications
  Applications Conference, SIU 2013}, pp.~1--4, IEEE, apr 2013.

\bibitem{Gupta2016}
O.~Gupta, D.~Raviv, and R.~Raskar, ``{Deep video gesture recognition using
  illumination invariants},'' {\em Arxiv}, pp.~1--9, 2016.

\bibitem{Chen2017}
X.~Chen, X.~Lan, G.~Liang, J.~Liu, and N.~Zheng,
  ``{Pose-and-illumination-invariant face representation via a triplet-loss
  trained deep reconstruction model},'' {\em Multimedia Tools and
  Applications}, vol.~76, pp.~22043--22058, nov 2017.

\bibitem{Gross2008}
R.~Gross, I.~Matthews, J.~Cohn, T.~Kanade, and S.~Baker, ``{Multi-PIE},'' in
  {\em 2008 8th IEEE International Conference on Automatic Face {\&} Gesture
  Recognition}, pp.~1--8, IEEE, sep 2008.

\bibitem{Kingma2015a}
D.~P. Kingma and J.~L. Ba, ``{Adam},'' pp.~1--15, 2015.

\bibitem{Shi2016}
W.~Shi, J.~Caballero, F.~Huszar, J.~Totz, A.~P. Aitken, R.~Bishop, D.~Rueckert,
  and Z.~Wang, ``{Real-Time Single Image and Video Super-Resolution Using an
  Efficient Sub-Pixel Convolutional Neural Network},'' in {\em 2016 IEEE
  Conference on Computer Vision and Pattern Recognition (CVPR)},
  pp.~1874--1883, 2016.

\bibitem{Lundqvist1998}
D.~Lundqvist, A.~Flykt, and A.~{\"{O}}hman, ``{The Karolinska Directed
  Emotional Faces - KDEF CD ROM from Department of Clinical Neuroscience,
  Psycology section},'' {\em Karolinska Institutet}, pp.~3--5, 1998.

\bibitem{Lucey}
P.~Lucey, J.~F. Cohn, T.~Kanade, J.~Saragih, Z.~Ambadar, and I.~Matthews,
  ``{The Extended Cohn-Kanade Dataset (CK+): A complete dataset for action unit
  and emotion-specified expression},'' tech. rep.

\bibitem{Lyons1998a}
M.~Lyons, S.~Akamatsu, M.~Kamachi, and J.~Gyoba, ``{Coding facial expressions
  with Gabor wavelets},'' in {\em Proceedings - 3rd IEEE International
  Conference on Automatic Face and Gesture Recognition, FG 1998}, pp.~200--205,
  1998.

\bibitem{Wallhoff2006a}
F.~Wallhoff, B.~Schuller, M.~Hawellek, and G.~Rigoll, ``{Efficient recognition
  of authentic dynamic facial expressions on the feedtum database},'' in {\em
  2006 IEEE International Conference on Multimedia and Expo, ICME 2006 -
  Proceedings}, vol.~2006, pp.~493--496, IEEE, jul 2006.

\bibitem{Ruiz-Garcia2018b}
A.~Ruiz-Garcia, N.~Webb, V.~Palade, M.~Eastwood, and M.~Elshaw, ``{Deep
  Learning for Real Time Facial Expression Recognition in Social Robots},'' in
  {\em Proceedings of the International Conference on Neural Information
  Processing}, 2018.

\bibitem{He2015a}
K.~He, X.~Zhang, S.~Ren, and J.~Sun, ``{Deep Residual Learning for Image
  Recognition},'' {\em Arxiv.Org}, vol.~7, pp.~171--180, dec 2015.

\bibitem{Hewahi2012}
N.~M. Hewahi and A.~R.~M. Baraka, ``{Impact of Ethnic Group on Human Emotion
  Recognition Using Backpropagation Neural Network},'' {\em BRAIN. Broad
  Research in Artificial}, pp.~20--27, 2012.

\bibitem{Kan2014}
M.~Kan, S.~Shan, H.~Chang, and X.~Chen, ``{Stacked progressive auto-encoders
  (SPAE) for face recognition across poses},'' in {\em Proceedings of the IEEE
  Computer Society Conference on Computer Vision and Pattern Recognition},
  pp.~1883--1890, IEEE, jun 2014.

\bibitem{Ruiz-Garcia2016a}
A.~Ruiz-Garcia, M.~Elshaw, A.~Altahhan, and V.~Palade, ``{Emotion Recognition
  Using Facial Expression Images for a Robotic Companion},'' in {\em
  Engineering Applications of Neural Networks: 17th International Conference,
  EANN 2016, Aberdeen, UK, September 2-5, 2016, Proceedings}, pp.~79--93,
  Springer, Cham, 2016.

\end{thebibliography}

\end{document}